\documentclass[letterpaper, 10 pt, conference]{ieeeconf} 

\IEEEoverridecommandlockouts                            

\usepackage{amsmath} 
\usepackage{upgreek}
\usepackage{graphicx}
\usepackage{dsfont}
\usepackage{multirow}

\title{\LARGE \bf
Multi-Modal Planning on Regrasping for Stable Manipulation
}

\author{Jiaming Hu, Zhao Tang, and Henrik I. Christensen
\thanks{Contextual Robotics Institute, UC San Diego, La Jolla, CA 92093, USA}}

\begin{document}

\maketitle
\thispagestyle{empty}
\pagestyle{empty}

\begin{abstract}

Nowadays, a number of grasping algorithms ~\cite{c1,c2} have been proposed, that can predict a candidate of grasp poses, even for unseen objects. This enables a robotic manipulator to pick-and-place such objects. However, some of the predicted grasp poses to stably lift a target object may not be directly approachable due to workspace limitations. In such cases, the robot will need to re-grasp the desired object to enable successful grasping on it. This involves planning a sequence of continuous actions such as sliding, re-grasping, and transferring. To address this multi-modal problem, we propose a Markov-Decision Process-based multi-modal planner that can rearrange the object into a position suitable for stable manipulation. We demonstrate improved performance in both simulation and the real world for pick-and-place tasks.

\end{abstract}

\section{INTRODUCTION}%
\label{sec:intro}
Pick-and-place is a core robotic manipulation task, that involves the application of end-effector force or torque at specific contact points on an object in order to move it from one place in the environment to another.
Several recent studies, such as Contact GraspNet~\cite{c2} and
Dex-Net~\cite{c1, c18}, have developed methods to quickly estimate the location of grasping poses for unseen objects. Based on the confidence level of the estimation method, the robot can then select a grasp, plan arm motion to grasp the target object, and place it in the desired position.

Several existing approaches have demonstrated their ability to use parallel grippers to grasp unknown objects~\cite{c1,c2}. Parallel grippers are often preferred over multi-finger grippers because parallel grippers have lower degrees-of-freedom, resulting in less dimensional complexity. However, a parallel gripper's lack of contact points with the target object can result in unstable grasps when gravitational torque from the object exceeds the static torque caused by friction, as highlighted in \cite{c3}. Therefore, it's critical for a robot with parallel grippers to select a \textit{high-quality} grasp that prevents rotational slippage. Unfortunately, existing methods often fail to consider the rotational stability of the object while predicting possible grasps. Additionally, executing a high-quality grasp may not always be feasible for mobile robots such as Fetch~\cite{c4}. However, with support from a receptacle surface such as a table, certain manipulations such as sliding may still be stable with only lower-quality grasps. In these situations, the robot can first slide the object into a position where high-quality grasps become available, then re-grasp the object. This two step process is expected to drastically improve grasping stability for pick-and-place manipulation.

\begin{figure}[!ht]
    \center
    \includegraphics[width=0.5\textwidth]{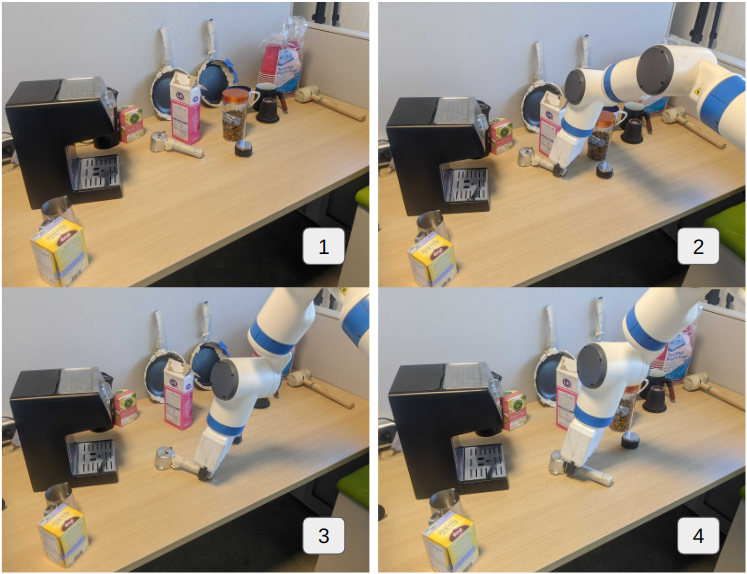}
    \caption{\label{main} (1)~Initially, the espresso machine handle is too close to the sugar box and a high-quality grasp is not feasible due to collision risk. \newline(2)~The robot grasps the edge of the handle and slides it to a better position. (3)~The handle is now in a position where a high-quality grasp is possible. (4)~The robot then lifts the object by using a stable grasp without collision.}
\end{figure}



This paper introduces a novel approach to enhance the stability of pick-and-place manipulation through sliding and re-grasping. To achieve this, we would need to differentiate between high and low-quality grasps and also be able to generate plans for re-grasping. The first aspect of our approach involves the development of a center of mass (COM) predictor, which enables us to classify grasp quality and identify reliable grasps minimizing slippage. In the second aspect, we propose a multi-modal planner to rearrange the object for better grasping by making high-quality grasp positions more accessible. By combining these two aspects, our system can significantly improve the stability of pick-and-place manipulation, both in simulation and real-world.

The outline of this paper is the following. In Section~\ref{sec:related}, related work and some terminology is introduced. We describe how we consider the manipulation problem a multi-modal question in Section~\ref{sec:problem}. We then show our methodology to construct and solve the problem in Section \ref{sec:method}. Subsequently, we will compare our methodology with previously published multi-modal planning methods in Section \ref{sec:comparsion}. Next, some implementation details are presented in Section \ref{sec:implementation}. Finally, Section~\ref{sec:experiments} demonstrates our work both in simulation and real-world.

\section{RELATED WORK}%
\label{sec:related}

\subsection{Learning-based Grasping Algorithms}
In the past few years, learning-based grasping algorithms have shown their ability to grasp unknown objects. For example, in 2018, Schmidt~\cite{c6} used a convolutional neural network (CNN) directly to predict a 6 DoF (degree of freedom) grasping pose from only a depth image. Later, in 2019, Liu~\cite{c7} proposed the work to predict multiple 6 DOF grasp poses by using a new loss function calculating the difference between the predicted pose and the closest ground truth grasp as the loss during training. In 2021, to reduce the difficulty of regressing all grasping poses, Sundermeyer~\cite{c2} proposes a method to use a 4 DoF representation for grasp poses instead of 6 DoF for a parallel gripper. Indeed, there are numerous extended efforts, such as refining grasp candidates~\cite{c8} or using reinforcement learning~\cite{c9}.

\subsection{Rearrangement Planning before Grasping}
When the target object is not in a position that is easy to grasp,
a re-arrangement of the environment will be required. One common non-graspable situation is for thin objects. To make it graspable, Hang~\cite{c10} proposed a solution to use the end-effector to press the object and move it to the table edge with
a slide so that the object becomes graspable. Another example is that when the target object is surrounded by other movable objects, Dogar~\cite{c11} proposed to slightly push objects to make room to grasp. Once the object is distant, 
 Daniel~\cite{c12} studied the sliding action for moving the object closer without considering obstacles. However, from our perspective, there is still no related work to slide and re-grasp unknown objects for better grasping in cluttered environments. Our planner has two key advantages over previous works. Firstly, it is capable of handling unknown objects. Secondly, in cases where a single rearrangement action is insufficient, our planner can generate plans that involve multiple actions to achieve the desired result.

\subsection{Multi-modal planning}
Multi-modal planning involves planning motion trajectories across different manifolds. Most~\cite{c13,c14} multi-modal planners have a two-level structure with two planners: a task planner, and a motion planner. The task planner is a high-level planner. It generates a sequence of manifolds to traverse through. The motion planner is a low-level planner. It generates the motion plan through the sequence manifolds provided by the task planner. Once the motion planner fails to plan in one manifold, the task planner will re-plan a new task sequence across manifolds until the motion planner solves all of them. In~\cite{c13}, Hauser proposed a multi-modal planner called Random-MMP, which built a tree to search through different manifolds. Later, because many multi-modal planning tasks have a foliation structure, Kingston~\cite{c14} proposed a mode transition graph to plan the path across manifolds. In~\cite{c14}, each node is a mode, while the edge contains a 2D weight graph representing the difficulty of switching between two states from two different modes. Then, they use the Dijkstra algorithm during planning to search for a path with the minimum total weight.

\section{PRELIMINARIES}%
\label{sec:prelim}
Because pick-and-place manipulation with rearrangement requires multiple continuous actions like sliding, re-grasping, and transferring, the problem has a multi-modal structure where its state space consists of a set of manifolds with differing constraints. Therefore, we will review some basic ideas to demonstrate how to solve a multi-modal problem, such as multi-modal motion planning and foliation.

\subsection{Multi-Modal Motion Planning}
In a multi-modal problem, the system is represented by a hybrid state $(q, \upvarsigma)$ where $q$ is the robot configuration in a continuous configuration space C and $\upvarsigma$ is a mode in a mode space $\Sigma$. Each mode $\upvarsigma$ defines a manifold (a configuration set) that satisfies certain constraints. A mode switch(or manifold transition) happens when there exists a feasible action to move between $(q_{a}, \upvarsigma_{a})$ and $(q_{b}, \upvarsigma_{b})$ where $\upvarsigma_{a}$ and $\upvarsigma_{b}$ are different manifolds. Then, the multi-modal problem asks for a feasible path connecting a start state $(q_{start}, \upvarsigma_{start})$ and a goal state $(q_{goal}, \upvarsigma_{goal})$.

\subsection{Foliation}
A foliation $F$ is a set of disjoint manifolds defined by a constraint function $f_{F}$ and co-parameter set $\{\theta_1, \theta_2, ... \}$. That is, each manifold $M_{\theta}$ of the foliation $F$ will be defined by one co-parameter $\theta$ with the constraint function $f_F$ of $F$ as shown in the following equation.
\[M_{\theta} = \{q \in C | f_{F}(q) = \theta \}\]

Manipulation problems, particularly those involving constrained motion planning, often have state spaces with a foliation structure. Consider the task of delivering a cup of water, where the constraint is that the cup must remain upright. However, depending on the way the cup is grasped, the constraint on the end-effector will differ, necessitating the use of multiple manifolds to describe the problem. To address this, the problem must be represented as a foliation, where each grasp becomes a co-parameter that defines a manifold for how the arm can maintain the cup's rotation with that particular grasp.

\section{PROBLEM FORMULATION}%
\label{sec:problem}
\begin{figure}[!ht]
    \center
    \includegraphics[width=0.49\textwidth]{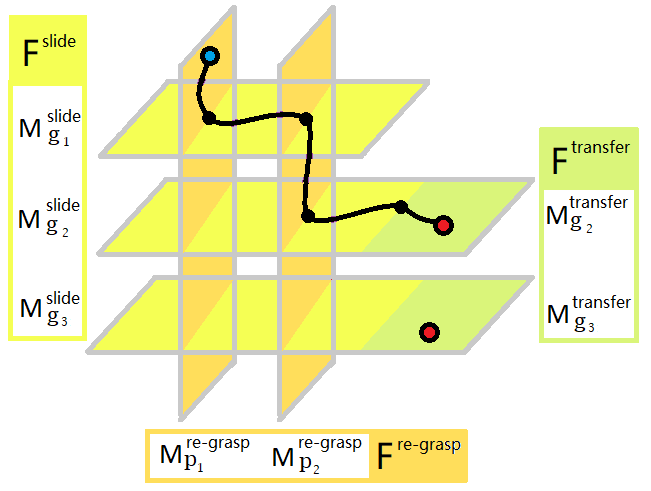}
    \caption{\label{foliation}$F^{re-grasp}$ are orange planes with its
      co-parameter set~($p_1$ and $p_2$). $F^{slide}$ are yellow planes with its co-parameter set~($g_1$, $g_2$, and $g_3$). $F^{transfer}$ are green planes with its co-parameter set~($g_2$ and $g_3$). The blue dot is the start state~($q_{start}$, $M^{re-grasp}_{p_1}$) with a start arm configuration $q_{start}$ in manifold $M^{re-grasp}_{p_1}$, while two red dots are two target states ($q_{target1}$, $M^{transfer}_{g_2}$) and ($q_{target2}$, $M^{transfer}_{g_3}$), where $q_{target1}$ and $q_{target2}$ are two target arm configurations in manifolds $M^{transfer}_{g_2}$ and $M^{transfer}_{g_3}$ respectively.}
\end{figure}

This paper addresses the problem of sliding and regrasping an unknown object for improved pick-and-place operations by conceptualizing it as a sequence of three foliations. The first foliation, denoted as $F^{slide}$, represents a family of manifolds where the robot grasps the target object, which can only slide horizontally while in contact with the table surface. The co-parameter of $F^{slide}$ consists of a set of grasp poses over the target object. The second foliation, denoted as $F^{transfer}$, represents manifolds where the robot is lifting and transferring the target object from one location to another, while maintaining a fixed vertical orientation, without any contact between the target object and the table. The co-parameter set in this case is a set of grasps that ensure stable lifting of the target object. Finally, the third foliation, $F^{re-grasp}$, represents a family of manifolds where the arm moves from one pre-grasp pose to another pre-grasp pose without touching both the environment and the target object. The co-parameter here is a set of object placements on the table surface. Fig.~\ref{foliation} illustrates the relationship between these foliations.

An intersection between two manifolds from different foliations exists when there are actions that enable the transition from a state in one manifold to another. In our work, we identify two types of intersections. The first intersection, between manifolds $M_{p}^{re-grasp}$ from $F^{re-grasp}$ and $M_{g}^{slide}$ from $F^{slide}$, encompasses all actions that involve releasing or grasping the target object in placement $p$ using grasp $g$. The second intersection, between manifolds $M_{g}^{transfer}$ from $F^{transfer}$ and $M_{g}^{slide}$ from $F^{slide}$, corresponds to the action of lifting the target object from the table surface using grasp $g$.

To formulate the problem, we consider the initial robot arm configuration $q_{init}$ and the current object placement $p_{start}$, along with the desired placement $p_{goal}$. We define the initial state as $(q_{init}, M_{p_{start}}^{re-grasp})$, where $M_{p_{start}}^{re-grasp}$ is the manifold of $F^{re-grasp}$ associated with the initial placement $p_{start}$. The set of target states comprises $(q_{target}, M_{g_{target}}^{transfer})$, where the robot can use the configuration $q_{target}$ to place the target object at $p_{goal}$ using grasp $g_{target}$. The problem addressed in this paper is finding a motion trajectory for the robot to cross different manifolds from the initial state to one of the target states, as illustrated in Fig.~\ref{foliation}.

\section{METHODOLOGY}%
\label{sec:method}

We follow a general pick-and-place pipeline, but our system includes rearrangement capabilities for situations where they are necessary. After it locates the table surface, the robot performs RGBD-based object segmentation to identify all present objects. Once the user selects the target object for grasping, the robot runs Contact GraspNet~\cite{c2} to identify a set of grasps to pick up and place the object. 

\begin{figure}[!ht]
    \center
    \includegraphics[width=0.5\textwidth]{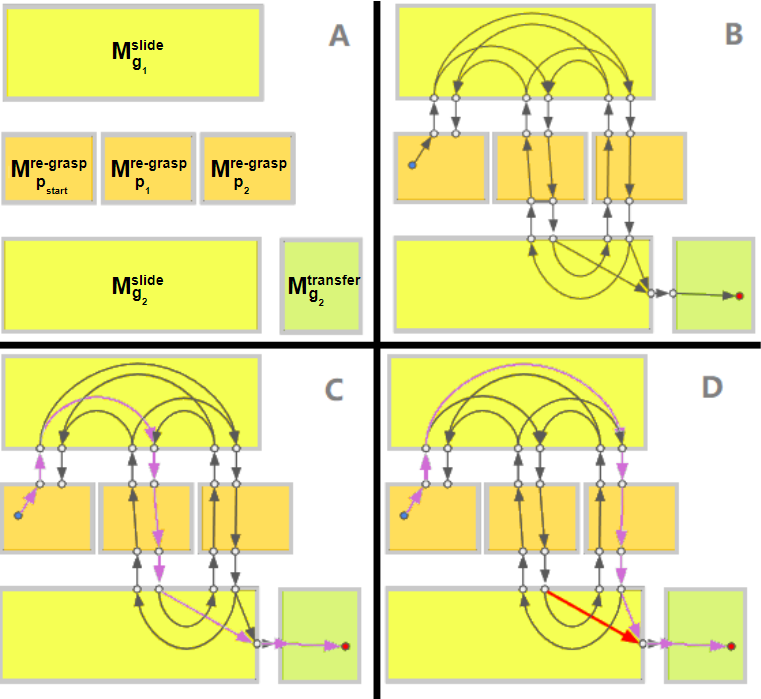}
    \caption{\label{main_pipeline} Four steps of the main pipeline: (A) Foliation Construction: Construct the foliation structure where the object has one low-quality grasp $g_1$ and one high-quality grasp $g_2$ along with three placements $p_1$, $p_2$, and $p_{start}$. (B) MDP Modeling: Convert the foliation structure into MDP format where each grey node is a state while edges are actions. The blue node is the initial state, while the red node is the target state. (C) Task and Motion Planning: The task planner plans a task sequence~(purple path) across different manifolds to ask the motion planner to solve. (D) Re-planning: The task planner re-plans a new task sequence~(purple path) if the motion planner fails to solve a task~(red directional edge) in the original task sequence. }
\end{figure}

However, our planner has a restriction that requires the robot to use a grasp that produces minimal gravitational torque during manipulation. To accomplish this, we have developed a COM predictor that estimates the target object's center of mass and determines which grasps are feasible based on the gravitational torque they would generate. If the robot cannot identify a feasible grasp that meets the low torque restriction, it initiates a rearrangement process. This process consists of four steps, which are illustrated in Fig.~\ref{main_pipeline}, and involves repositioning the target object to make it more suitable for manipulation.

The structure of this section is listed as the following:

\begin{enumerate}
  \item Grasp Analysis: Identify both high-quality and low-quality grasps based on their poses and the target object's COM estimated by the COM predictor.
  \item Foliation Construction: Convert the pick-and-place with rearrangement problem into a foliation structure.
  \item Markov-Decision Process Modeling: Model the foliation structure as a Markov-Decision Process(MDP).
  \item Task and Motion Planning: Plan a sequence of motion tasks cross through different manifolds based on the generated MDP, and solve those motion tasks to generate the robot motion trajectory for accomplishing the manipulation.
  \item Re-planning: Re-plans a task sequence until all motion tasks in the task sequence are solved. 
\end{enumerate}

\subsection{Grasp Analysis}

\begin{figure}[!ht]
    \center
    \includegraphics[width=0.5\textwidth]{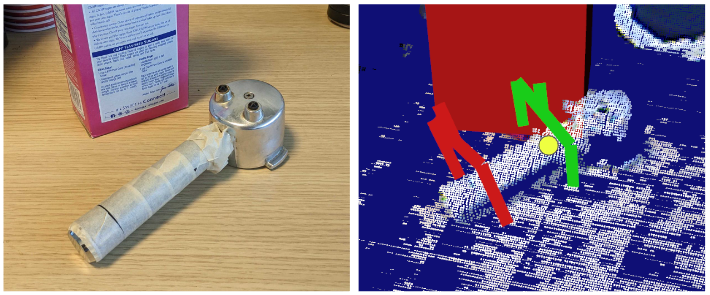}
    \caption{\label{grasp_analysis} Left: the RGB image of the espresso machine handle. Right: The yellow dot is the predicted COM of the handle. The green graphics is for the grasp with low gravitational torque, while the red graphics is for the grasp with high gravitational torque.}
\end{figure}

Our grasp analysis process involves two main steps. First, we use Contact GraspNet~\cite{c2} to predict a set of grasps over the target object, based on the entire point cloud of the scene and its segmented point cloud. Next, we classify these grasps as either high-quality or low-quality, based on their stability in lifting the object. To do this, we use a COM predictor, with PointNet++~\cite{c21} as backbone, trained by ACRONYM~\cite{c5}, which estimates the target object's COM by analyzing its segmented point cloud and an imaginary table point cloud. We then compare the predicted COM with the current grasp pose to rank the grasps in terms of their gravitational torque. Grasps with lower gravitational torque are considered high-quality, while those with higher gravitational torque are deemed low-quality. This analysis allows us to select the stable grasps for lifting the target object. An example of the classification result is shown in Fig.~\ref{grasp_analysis}. In the image, the green grasp is classified as high-quality because the horizontal distance between it and the COM is short, resulting in a small gravitational torque. Conversely, the red grasp is classified as low-quality because it produces a higher gravitational torque.

\subsection{Foliation Construction}

Foliation construction requires the co-parameter sets for defining each foliation. For example, for the $F^{re-grasp}$ foliation, we require a set of collision-free placements on the table surface as the co-parameter set. Similarly, for the $F^{transfer}$ and $F^{slide}$ foliations, we need a set of grasps as the co-parameter set.

\subsubsection{$F^{re-grasp}$ construction}

After segmenting the target object, we sample a set of collision-free placements $\{p_1, p_2, \ldots\}$ on the table surface with other obstacles, using the target object's bounding box as a reference. Each $p_i$ represents defines a manifold $M_{p_i}^{re-grasp}$ of $F^{re-grasp}$.

\subsubsection{$F^{transfer}$ and $F^{slide}$ construction}

Each high-quality grasp $g_h$ with low gravitational torque defines both manifold $M_{g_h}^{transfer}$ and manifold $M_{g_h}^{slide}$. On the other hand, each low-quality grasp $g_l$ with high gravitational torque defines a manifold $M_{g_l}^{slide}$.

\subsection{Modeling as a Markov-Decision Process}

\begin{figure}[!ht]
    \center
    \includegraphics[width=0.45\textwidth]{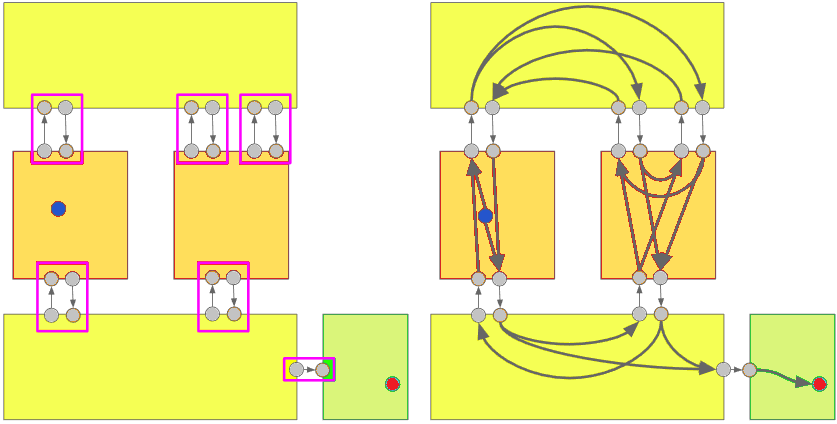}
    \caption{\label{mdp_foliation} Left: Inter-manifold actions involve the interplay between $M_{g}^{slide}$ and $M_{p}^{re-grasp}$ manifolds, where each red box contains approach and release actions, computed via a Cartesian motion planner, which are essentially mirror trajectories. For transitions between $M_{g}^{slide}$ and $M_{g}^{transfer}$, a solitary lifting action, in the red box, raises the object. Right: Intra-manifold actions are represented by arrows within each manifold: re-grasping in $M_{p}^{re-grasp}$, sliding in $M_{g}^{slide}$, and transferring post-pickup in $M_{g}^{transfer}$.}
\end{figure}

To model the foliation structure of the problem as an MDP, we need to define the state, action, transition probability, and reward. First, we will explain how to define the actions and states. Then, we will describe how we define the transition probability and reward.

\subsubsection{Action and State}

To define actions and states, we first need to search for actions as intersections between two manifolds from different foliations, as illustrated on the left side of Fig.~\ref{mdp_foliation}. In our generated foliation structure, there are only two types of intersections: one between $M_{g}^{slide}$ and $M_{p}^{re-grasp}$, and the other between $M_{g}^{slide}$ and $M_{g}^{transfer}$ with the same co-parameter g. Because the robot can't lift the target object without touching it, intersections between $M_{p}^{re-grasp}$ and $M_{g}^{transfer}$ do not exist.

The intersection between $M_{g}^{slide}$ and $M_{p}^{re-grasp}$ are actions to approach and release the target object with grasp g at placement p, as shown in the left of Fig.~\ref{mdp_foliation}. Releasing the target object with grasp g at placement p is an action from $M_{g}^{slide}$ to $M_{p}^{re-grasp}$. Conversely, approaching the target object with grasp g at placement p is an action from $M_{p}^{re-grasp}$ to $M_{g}^{slide}$. To find these actions, we sample multiple arm configurations $\{q_1, q_2, ..., q_i, ... \}$ using inverse kinematics (IK) solver to grasp the target object at placement p with grasp g. We plan a releasing motion trajectory for each arm configuration $q_i$ to a pre-grasp pose of g using a Cartesian motion planner~\cite{c20}. Therefore, this releasing motion trajectory becomes one of the releasing actions from $M_{g}^{slide}$ to $M_{p}^{re-grasp}$. Similarly, the reverse of this trajectory represents the approaching action from $M_{p}^{re-grasp}$ to $M_{g}^{slide}$.

The intersection between $M_{g}^{slide}$ and $M_{g}^{transfer}$ are the actions to lift the target object with high-quality grasp, and we can find them by reusing the sampled arm configurations $\{q_1, q_2, ..., q_i, ... \}$. Given a high-quality grasp $g_h$, from the previous sampled arm configuration set, we find all arm configurations $\{q_1, q_2, ..., q_j, ...\}$ which can grasp the target object with grasp $g_h$. Then, for each $q_j$, we use the Cartesian motion planner~\cite{c20} to plan the lifting motion trajectory, and this motion trajectory will be a lifting action from $M_{g}^{slide}$ to $M_{g}^{transfer}$.

We now have a collection of sampled actions with motion trajectories between manifolds. Each of these trajectories has an end arm configuration, represented as gray nodes in Fig.~\ref{mdp_foliation}. We can consider these end configurations as states of MDP.

Next, we focus on searching for actions within a single manifold, as depicted in the right-hand side of Fig.~\ref{mdp_foliation}. In each $M_{g}^{slide}$, all incoming states have an action connecting to all outgoing states as long as both states are not between the same pairs of manifolds. In each $M_{p}^{re-grasp}$, all incoming states have an action connecting to all outgoing states as long as they don't have the same arm configuration. Finally, in each $M_{g}^{transfer}$, all incoming states have an action connecting to all target states in the same manifold.

Eventually, the current arm configuration $q_{init}$ serves as the initial state $(q_{init}, M_{p_{start}}^{re-grasp})$ where $p_{start}$ denotes the current placement of the target object. Furthermore, this initial state has actions connecting to all outgoing states in $M_{p_{start}}^{re-grasp}$. Additionally, for each manifold $M_{g_h}^{transfer}$, all arm configurations $q_{target}$ placing the target object to the desired position are the target state $(q_{target}, M_{g_h}^{transfer})$.

\subsubsection{Transition Probability and Reward}

Each action $a$ has its own success possibility $\rho_a$ used to define the transition probability. The success possibility $\rho_a$ represents the probability of finding a feasible motion trajectory from the current state to the next state. For example, given an action $a$ from the state $S_{current}$ to state $S_{next}$, the transition probability is that the new state will be $S_{next}$ with a possibility $\rho_a$. At the same time, it may lead to a failure state, a terminal state, with a possibility $1-\rho_a$. Initially, all actions in the manifold have a 0.5 success possibility, while all actions between manifolds have a 1.0 success probability because they have motion trajectories already.

The reward function used in this approach takes only the state as input. Only the target states, which are terminal states, have a positive reward. On the other hand, all other states have a small negative reward, and the failure state (also a terminal state) has a large negative reward.

\subsection{Task and Motion Planning}

The objective of task and motion planning is to generate a path from the initial configuration to one of the target configurations across multiple manifolds, using two planners. The first planner is a task planner, which generates a sequence of tasks. Each task corresponds to a motion planning task with some constraints. The second planner is a motion planner, which solves each constrained motion planning task in the task sequence generated by the task planner. Once the motion planner has solved all tasks and returned a list of robot motion trajectories, the accumulated motion trajectory is considered the final result.

\subsubsection{Task planner}

To generate a task sequence, the task planner applies value iteration. Each state in the MDP contains a value, which is updated by the following equation, except for terminal states.

\[v_{k+1}=\max_{a} \sum_{s'}p(s'|s,a)[R(s') + \gamma v_k (s')]\]
\[=\max_{a}[p_{success}(R(S_a) + \gamma v_k (S_a)) + p_{fail}R(S_{failure})]\]
where $S_a$ is the state which action $a$ should lead to from state $S$, $R(S_a)$ is the reward of entering state $S_a$, $\gamma$ is a discount factor, $p_{success}$ is the success possibility $\rho_a$ of action $a$, $p_{fail}$ is $1-\rho_a$, and $R(S_{failure})$ is the high penalty to the failure state.

After performing value iteration, the optimal action for each state in the MDP structure is determined, which is the action that leads to the neighboring state with the highest value. Using this information, the task planner generates an action sequence by continuously selecting the optimal action from the initial state until the target state is reached. After discarding all actions between manifolds, the resulting action sequence is represented as $[a_1, a_2, ...]$, where each action $a_i$ is a directional edge in one manifold $M_{a_i}$, connecting from an arm configuration $q_{start}^{a_i}$ to another arm configuration $q_{end}^{a_i}$. Consequently, the task sequence is a list $[t_1, t_2, ...]$ of motion planning tasks, where each $t_i$ corresponds to a motion planning task from $q_{start}^{a_i}$ to $q_{end}^{a_i}$ with the constraints of $M_{a_i}$.

\subsection{Motion Planner}

Given the generated task sequence, the motion planner must utilize a constrained motion planning algorithm to solve each motion task with certain constraints. However, in scenarios where re-planning may be necessary, repeatedly planning within a single manifold may be required. For this reason, we decide to use the lazy probabilistic roadmap (LazyPRM) planner~\cite{c15}, offering an efficient approach for multi-query situations. Nonetheless, the original LazyPRM is unsuitable for constrained motion planning by default. Therefore, we propose a modified version of the LazyPRM that incorporates the projection method\cite{c17} to become a constrained-based LazyPRM. As a result, this augmented planner can effectively account for the constraints present in each manifold and generate a trajectory that satisfies them.


Since each manifold has unique constraints, the constrained-based LazyPRM algorithm must carefully consider these constraints when planning trajectories. For example, when planning in the manifold $M_{g}^{slide}$, the object is grasped in grasp $g$ and becomes the new end-effector of the arm, so the motion task has a constraint that restricts the object to move only within the plane of the table surface. Furthermore, in $M_{g}^{transfer}$, the motion task has a vertical rotation constraint focusing on maintaining the consistency of the horizontal distance between the grasping point and the center of mass to avoid slippage. Finally, in $M_{p}^{re-grasp}$, the motion task has no constraints but still considers the object as an obstacle in placement p during planning. 

\subsection{Re-planning}

If the motion planner fails to find a solution for a task in the sequence, it promptly informs the task planner of the failure. The task planner then updates the transition probability and re-plans the task sequence. Assuming the task sequence is represented as ${t_1, t_2, \ldots, t_f, \ldots}$ with $t_f$ being the task that failed to find a solution, we have the action sequence ${a_1, a_2, \ldots, a_f, \ldots}$ where each action corresponds to a task in the sequence. In the MDP, the task planner can set the success possibility of all actions before $a_f$ to 1.0 and decrease the success possibility of $a_f$. However, with this updated policy, the task planner will re-plan a task sequence similar to the previous one. To prevent this, we can reduce the success possibility of all similar actions to $a_f$. For example, if an action slides the object from position $p_a$ to position $p_b$ with grasp g, we can reduce the success possibility of all actions that use a different grasp to slide the object from $p_a$ to $p_b$. Our experiments strongly suggest that this new policy can significantly improve run-time, similar to the experiment in \cite{c14}. Moreover, during the success possibility update, our task planner will not reduce the success possibility of actions that already have a solution.

\section{Comparison with Mode Transition Graph}
\label{sec:comparsion}

Our MDP-based approach offers several advantages over the multi-modal planner described in~\cite{c14}, which relies on a mode transition graph. This graph assigns weights to state transitions based on their perceived difficulty, with lower weights indicating easier transitions. During task sequence planning, their task planner utilizes the Dijkstra algorithm to identify the path with the lowest cumulative weight in the mode transition graph. If the motion planner successfully solves a transition, the task planner applies a small penalty to its edge weight. However, if a transition cannot be solved, a larger penalty is applied to its edge weight to discourage future attempts. We list two major advantages over this approach.

Firstly, our approach offers more consistent metrics to guide the motion planning process. Selecting the correct parameters for weight adjustment in state transition can be challenging if the weights are defined arbitrarily, which may lead to inconsistent weight adjustments that confuse the planner. For instance, in the current approach, while multiple transitions all have solutions, they may have different weights. On the other hand, our approach uses probability to update a value function, leading to a more rigorous and thus more consistent heuristic. For instance, given the same situation of multiple transitions all having solutions, our method would assign a probability of 1.0 indicating that a solution exists. 


Secondly, by using a transition probability and reward-based system to guide our planner, we introduce additional layers of flexibility. For instance, if we want to discourage the robot from entering certain modes based on prior knowledge, such as placement poses that are far way from the robot, we can assign nodes in such modes negative rewards. This is not possible to do with a mode transition graph.


Thirdly, our task planner offers a more logically re-planning process with its correction mechanism. In the original approach described in~\cite{c14}, when the motion planner fails to solve a transition, the task planner applies a higher penalty to the weight of all “nearby” transitions in the mode transition graph. However, this approach may lead to unnecessary penalties on transitions that already have a solution. In contrast, our proposed approach utilizes probabilities to measure the difficulty of mode transitions, which enables us to avoid such penalties. Specifically, if a transition has a success probability of 1.0, it is deemed to be a solved transition, and the task planner will not apply any additional penalty to it.

Fourthly, our task planner has a more efficient method of reusing the transitions that have solutions. In contrast to the mode transition graph approach used in~\cite{c14}, which applies a small penalty to the weight of a transition that the motion planner successfully solves, we have found this approach to be unsuitable for constrained motion planning, where finding additional plans may be too costly. Instead, our approach increases the success rate of a transition once it has been solved, thereby increasing its expected value and encouraging the task planner to reuse transitions that have already been solved. This method leads to increased efficiency in generating a robot motion trajectory across multiple manifolds and allows for more efficient use of resources than the approach used in~\cite{c14}.

\section{IMPLEMENTATION DETAILS}%
\label{sec:implementation}
This section will discuss some challenges we met during the experiment, and the solutions to them.

\subsection{Uncertain Object Motion during Grasping}

During experiments, we realized that each grasping action might cause uncertainty in the in-hand object pose. That is, the grasps generated from the
Contact GraspNet have no guarantee of the consistency of the object pose in hand
during grasping. As a result, after grasping and sliding the object, the object
may not be in the pose that the robot expects, so the re-grasping action will
fail quickly due to the wrong pose estimation. Therefore, during the experiment,
after grasping and sliding the object, the robot will move the arm away from the
camera view and identify the target object. Thus, it can rerun the Contact
GraspNet and re-plan the rearrangement. Once the robot realizes there is no need
to rearrange the object for pick-and-place, then the robot can lift the object
directly with a stable grasp.

\subsection{Occlusion by other obstacles}

During manipulation, the robot will run both Contact GraspNet and COM predictor
to read the segmented object point cloud and predict both grasps and the target
object's com. However, both predictors may return the wrong estimation when
other objects occlude the target object, so it may cause manipulation failure.
To overcome this, we can find the object occluding the target object based on
the point cloud, similar to~\cite{c16} and run the grasping algorithm
recursively. Eventually, all objects occluding the target object will be
removed, so both predictors can have a better result on the target object.

\section{EXPERIMENTAL EVALUATION}%
\label{sec:experiments}

To demonstrate the improved stability of our planner, we compared it to a baseline planner that does not consider rearrangement steps involving sliding and re-grasping. Specifically, once the object to be grasped is selected, the robot utilizes Contact GraspNet to predict grasps and selects one of them for the pick-and-place manipulation. By conducting this comparison, we aim to highlight the benefits of incorporating rearrangement steps into the planning process. Thus, this section demonstrates the performances of our planner in the pick-and-place task with Fetch robot \cite{c4} in both simulation and the real world compared to the baseline, which does not consider rearrangement.

\subsection{Simulation Experiment}
\subsubsection{Setup}
We conducted simulation experiments using 12 different objects in 12 distinct scenes in Coppeliasim with the Bullet physics engine. The scenes comprised multiple objects placed on a table. A Fetch~\cite{c4} platform, having a parallel gripper with 100 N grasping force, is positioned at some distance in front of the table. In the first 6 scenes, the target object was positioned in a location that made it difficult to achieve high-quality grasps, either due to collisions with nearby objects or being out of range. For the remaining 6 scenes, the target objects were placed randomly on the table as long as they were collision-free.

\subsubsection{Experiment Trials and Stability Metric}
A trial in the experiment involved the robot grasping the object and then removing the supporting table from the simulation. After the table has been removed for 5 seconds, a stability score is measured. \emph{Grasping Stability} is measured by the inverse of the object's rotational change relative to the gripper after the table has been removed: $\textrm{stability} = \frac{1}{\Delta R}$.  

\begin{table}[!h]
\caption{Simulation Result}
\label{sim_result}
\begin{center}
\begin{tabular}{|c|c|cc|}
\hline
\multirow{2}{*}{object name} & \multirow{2}{*}{object mass(kg)} & \multicolumn{2}{c|}{stability (higher is better)}             \\ \cline{3-4} 
                             &                              & \multicolumn{1}{c|}{baseline} & our method \\ \hline
hammer                       & 0.10                          & \multicolumn{1}{c|}{1.04}        & \textbf{9.06}         \\ \hline
pan                          & 0.16                          & \multicolumn{1}{c|}{1.71}        & \textbf{5.91}         \\ \hline
candy bar                       & 0.16                          & \multicolumn{1}{c|}{0.00}        & \textbf{32.6}         \\ \hline
microphone                     & 0.10                          & \multicolumn{1}{c|}{23.5}        & \textbf{41.6}         \\ \hline
level                      & 0.51                          & \multicolumn{1}{c|}{0.00}        & \textbf{6.23}         \\ \hline
wrench                           & 0.12                          & \multicolumn{1}{c|}{0.00}        & \textbf{13.5}         \\ \hline
tissue box                      & 0.80                          & \multicolumn{1}{c|}{3.92}        & \textbf{4.36}         \\ \hline
cereal                       & 0.12                          & \multicolumn{1}{c|}{5.94}        & \textbf{25.3}         \\ \hline
can                          & 0.10                          & \multicolumn{1}{c|}{19.3}        & \textbf{68.6}         \\ \hline
caliper                     & 0.18                          & \multicolumn{1}{c|}{5.06}        & \textbf{8.29}         \\ \hline
dispenser                     & 0.25                          & \multicolumn{1}{c|}{0.71}        & \textbf{35.9}         \\ \hline
remote                     & 0.23                          & \multicolumn{1}{c|}{0.35}        & \textbf{5.11}         \\ \hline

\end{tabular}
\end{center}
\end{table}

\subsubsection{Results}
Results are shown in Table \ref{sim_result}. In all of the scenes, our multi-modal task planner achieved on-par or higher stability scores compared to the baselines. For certain scenes and objects, the baseline was unable to complete the grasping task entirely, resulting in a stability score of 0. In contrast, our multi-modal planner were able to execute stable grasps in all of the scenes. This proves that the multi-modal planner is more flexible and capable of manipulating objects in a much stabler fashion compared to the uni-modal baseline planner.

\subsection{Real-world Experiments}

\begin{figure}[!ht]
    \center
    \includegraphics[width=0.49\textwidth]{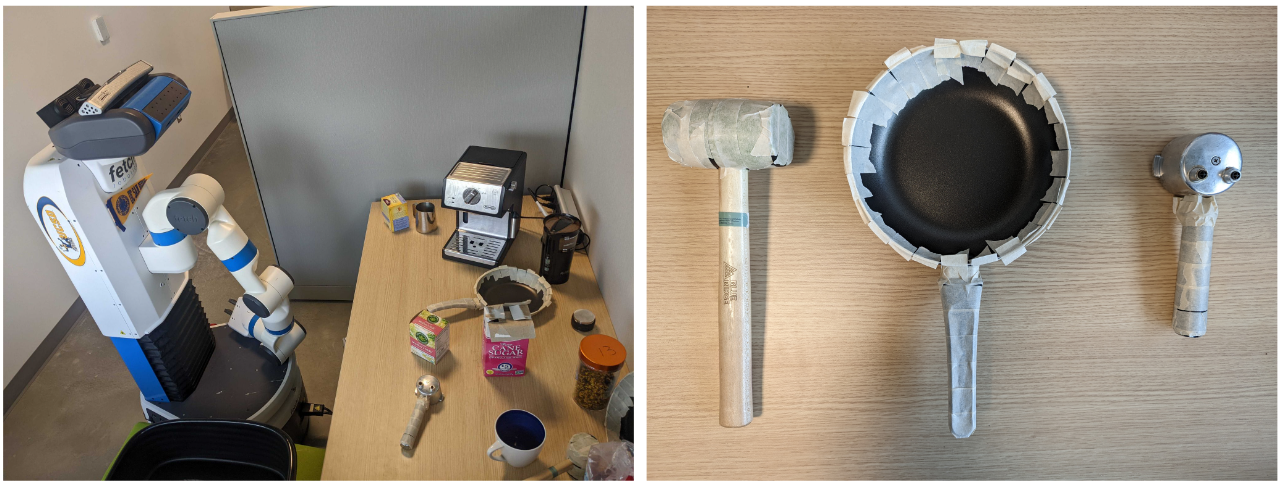}
    \caption{\label{manipulation_objects} Left: the setup of the real-world experiment with Fetch. Right: Objects~(hammer, pan, and espresso machine handle) used in the real-world experiment.}
\end{figure}

\subsubsection{Setup}

We conducted real-world experiments on three objects - a hammer, a pan, and an espresso machine handle - placed on a cluttered table, as shown in Fig.~\ref{manipulation_objects}. Initially, the hammer and pan were positioned in a pose where achieving high-quality grasps was not feasible, while the espresso machine handle was placed near an obstacle that made high-quality grasps unattainable due to collision. During running the experiment, we tasked the robot with moving those objects into a pre-placed bin with both the baseline and multi-modal planners.

\subsubsection{Results} 
\begin{figure}[!ht]
    \center
    \includegraphics[width=0.37\textwidth]{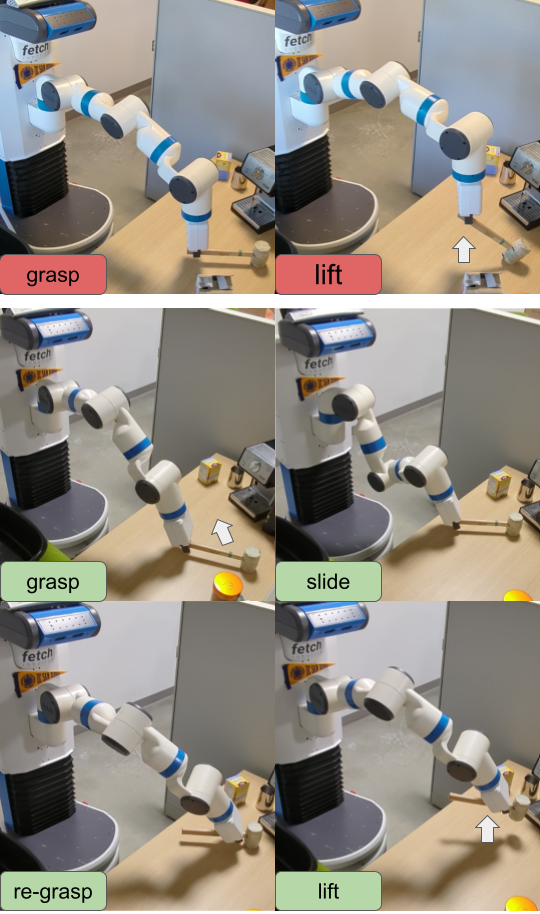}
    \caption{\label{real_world_hammer} Upper: Without rearrangement, the hammer slips around the grasping point during lifting. Lower: With rearrangement, before lifting, the robot slides the hammer into a better position for reliable grasping.}
\end{figure}

Our multi-modal planner was successful in placing objects into the bin after a few trials. The baseline planner, on the other hand, was unable to successfully place the objects entirely.  An example of a trial is shown in Fig.~\ref{real_world_hammer}. The hammer was initially positioned in a way that only allowed the robot to grasp its handle. With the baseline planner, the upper two figures illustrate a scenario without using rearrangement. In this case, once the gripper grasped the handle and lifted it up, the gravitational torque caused the hammer to slide down and remain in contact with the table. On the other hand, as shown in the lower four figures, our planner determined that the low-quality grasps on the handle were insufficient to lift the hammer, prompting the planner to rearrange the hammer to bring it closer. This adjustment enabled the robot to grasp the hammer in the middle where the predicted center of mass (COM) caused minimal gravitational torque and lift it up successfully.

\subsubsection{Failure Cases}
The most common failure case for the multi-modal planner was sparse grasping positions generated by Contact Graspnet, resulting in it being unable to find a solution. The most common failure cases for the baseline planner were 1. the planner failed to find a placement plan due to scene clutter 2. the object slipped or swung after grasping and resulted in a collision with the scene or the gripper dropping the object.

\section{CONCLUSIONS}%
\label{sec:conclusion}
Grasping previously unseen objects is by now within the realm of everyday robotics. However, it may not always be possible to directly generate a stable grasp for manipulation due to workspace limitations. To address this challenge, we propose a methodology to analyze a particular scenario to determine if a stable grasp is directly available. If not, we propose a restructuring motion to move the object to a position, from which a stable grasp for manipulation can be attained. We organize the process into a sequence of three foliations, i.e., a set of manifolds. A Markov decision process is used for sequencing the manifolds, and a geometric motion planner is used for trajectory planning. The performance of the new approach is demonstrated both in simulation with significant improvement in grasp stability, and we also demonstrate the performance for a set of real-world objects manipulated by a Fetch manipulator. The new method significantly expands the ability for a robot to pick up and manipulate objects in everyday cluttered environments. 

\addtolength{\textheight}{-12cm}%



\begin{thebibliography}{99}

  \bibitem{c1} J. Mahler, J. Liang, S. Niyaz, M. Laskey, R. Doan, X. Liu, J. A.
  Ojea, and K. Goldberg, ``Dex-net 2.0: Deep learning to plan robust grasps
  with synthetic point clouds and analytic grasp metrics.'' Robotics Science and
  Systems, (2017).

  \bibitem{c2} M. Sundermeyer, A. Mousavian, R. Triebel, and D. Fox.
  ``Contact-GraspNet: Efficient 6-DoF Grasp Generation in Cluttered Scenes.'' In
  2021 IEEE International Conference on Robotics and Automation (ICRA). IEEE
  Press, 13438–13444. (2021) https://doi.org/10.1109/ICRA48506.2021.9561877

  \bibitem{c3} K. Tsuchiya, S. Kagami, W. Yoshizaki and H. Mizoguchi, ``Grasp
  planning pre-computation by considering center of gravity of objects and its
  evaluation using OpenRAVE.'' 2012 IEEE International Conference on Systems,
  Man, and Cybernetics (SMC), Seoul, Korea (South), 2012, pp. 2091-2096, doi:
  10.1109/ICSMC.2012.6378048.

  \bibitem{c4} M. Wise, M. Ferguson, D. King, E. Diehr and D. Dymesich, ``Fetch
  and Freight: Standard Platforms for Service Robot Applications.'', Workshop on
  Autonomous Mobile Service Robots (2016), pp. 1--6

  \bibitem{c5} C. Eppner, A. Mousavian and D. Fox, ``ACRONYM: A Large-Scale Grasp
  Dataset Based on Simulation,'' 2021 IEEE International Conference on Robotics
  and Automation (ICRA), Xi'an, China, 2021, pp. 6222-6227, doi:
  10.1109/ICRA48506.2021.9560844.

  \bibitem{c6} P. Schmidt, N. Vahrenkamp, M. Wächter and T. Asfour, ``Grasping of
  Unknown Objects Using Deep Convolutional Neural Networks Based on Depth
  Images,'' 2018 IEEE International Conference on Robotics and Automation (ICRA),
  Brisbane, QLD, Australia, 2018, pp. 6831-6838, doi: 10.1109/ICRA.2018.8463204.

  \bibitem{c7} M. Liu, Z. Pan, K. Xu, K. Ganguly, and  D. Manocha, 
  ``Generating Grasp Poses for a High-DOF Gripper Using Neural Networks''. 2019
  IEEE/RSJ International Conference on Intelligent Robots and Systems (IROS),
  (2019) 1518-1525.

  \bibitem{c8} Y. Zhou and K. Hauser. ``6-DOF grasp planning by optimizing
  a deep learning scoring function.'' Robotics: Science and systems (RSS)
  workshop on revisiting contact-turning a problem into a solution. Vol. 2.
  (2017).

  \bibitem{c9} S. Levine, P. Pastor, A. Krizhevsky, and D. Quillen, ``Learning
  hand-eye coordination for robotic grasping with deep learning and large-scale
  data collection.'' The International journal of robotics research 37.4-5
  (2018): 421-436.

  \bibitem{c10} K. Hang, A. S. Morgan and A. M. Dollar, ``Pre-Grasp Sliding
  Manipulation of Thin Objects Using Soft, Compliant, or Underactuated Hands,''
  in IEEE Robotics and Automation Letters, vol. 4, no. 2, pp. 662-669, April
  2019, doi: 10.1109/LRA.2019.2892591.

  \bibitem{c11} M. R. Dogar and S. S. Srinivasa. ``Push-grasping
  with dexterous hands: Mechanics and a method.'' 2010 IEEE/RSJ International
  Conference on Intelligent Robots and Systems. IEEE, 2010.

  \bibitem{c12} D. Kappler, L Y. Chang, N. Pollard, T. Asfour, and R. Dillmann,
  ``Templates for pre-grasp sliding interactions.'' Robotics and Autonomous
  Systems 60.3 (2012): 411-423.

  \bibitem{c13} K. Hauser and V. Ng-Thow-Hing ``Randomized multi-modal motion
  planning for a humanoid robot manipulation tas''. The International Journal of
  Robotics Research. (2011); 30(6):678-698. doi:10.1177/0278364910386985

  \bibitem{c14} Z. Kingston, A. M. Wells, M. Moll and L. E. Kavraki, ``Informing
  Multi-Modal Planning with Synergistic Discrete Leads,'' 2020 IEEE International
  Conference on Robotics and Automation (ICRA), Paris, France, 2020, pp.
  3199-3205, doi: 10.1109/ICRA40945.2020.9197545.

  \bibitem{c15} R. Bohlin and L. E. Kavraki, ``Path planning using lazy PRM,''
  Proceedings 2000 ICRA. Millennium Conference. IEEE International Conference on
  Robotics and Automation. Symposia Proceedings (Cat. No.00CH37065), San
  Francisco, CA, USA, 2000, pp. 521-528 vol.1, doi: 10.1109/ROBOT.2000.844107.

  \bibitem{c16} A. Murali, A. Mousavian, C. Eppner, C. Paxton and D. Fox,
    ``6-DOF grasping for target-driven object manipulation in clutter.'' 2020
    IEEE International Conference on Robotics and Automation. IEEE, 2020.

  \bibitem{c17} D. Berenson, S. S. Srinivasa, D. Ferguson and J. J. Kuffner, ``Manipulation planning on constraint manifolds,'' 2009 IEEE International Conference on Robotics and Automation, Kobe, Japan, 2009, pp. 625-632, doi: 10.1109/ROBOT.2009.5152399.

  \bibitem{c18} J. Mahler et al., ``Dex-Net 1.0: A cloud-based network of 3D objects for robust grasp planning using a Multi-Armed Bandit model with correlated rewards,'' 2016 IEEE International Conference on Robotics and Automation (ICRA), Stockholm, Sweden, 2016, pp. 1957-1964, doi: 10.1109/ICRA.2016.7487342.

  \bibitem{c19} J. Hu and H. I. Christensen, ``Rotational Slippage Minimization in Object Manipulation,'' 2022 IEEE 18th International Conference on Automation Science and Engineering (CASE), Mexico City, Mexico, 2022, pp. 1896-1903, doi: 10.1109/CASE49997.2022.9926618.

  \bibitem{c20} D. Coleman, I. A. Șucan, S. Chitta, N. Correll, ``Reducing the Barrier to Entry of Complex Robotic Software: a MoveIt! Case Study'', Journal of Software Engineering for Robotics, 5(1):3–16, May 2014.

\bibitem{c21} Q. Charles R, Y. Li, S. Hao, G. Leonidas J, ``PointNet++: Deep Hierarchical Feature Learning on Point Sets in a Metric Space'', arXiv preprint arXiv:1706.02413, 2017.

\end{thebibliography}
\end{document}